%% file: main.tex
\documentclass[manuscript,nonacm]{acmart}

\AtBeginDocument{%
  }

\usepackage{longtable}
\usepackage{array}
\usepackage{calc}
\usepackage{booktabs}

\title[Adding One More Agent]{Adding One More Agent: How AI Team Size and Formation Shape Productivity and Diversity in Human-AI Brainstorming Teams}
\author{Rong-Ching Chang}
\affiliation{%
  \institution{University of California, Davis}
  \city{Davis}
  \state{California}
  \country{USA}}
\email{rocchang@ucdavis.edu}

\author{Hao-Chuan Wang}
\affiliation{%
   \institution{University of California, Davis}
  \city{Davis}
  \state{California}
  \country{USA}}
\email{hciwang@ucdavis.edu}
\renewcommand{\shortauthors}{Chang et al.}

\begin{document}


\begin{abstract}
Generative AI has improved individual productivity, but whether adding a plurality of AI partners helps or hurts remains open. We crossed AI partnership (one versus two) with behavioral formation (calibrated to average human behavior versus customized by the participant) in a brainstorming experiment ($N$=168) with human-human dyads as reference. Two-AI rooms posted more ideas to the shared board, but human-authored ideas dropped substantially. With one AI partner, formation mattered: humans paired with an average-calibrated AI posted more than any other condition. With two, this effect vanished. Humans continued talking but converted fewer ideas into board posts. AI social talk predicted more human posts, while AI posting volume predicted fewer. AI presence increased global diversity of human-authored ideas, but only a second AI increased within-session differentiation. The number of AI partners and their formation can reshape what the human contributes; single-agent results do not generalize to multi-agent teams.
\end{abstract}

\begin{teaserfigure}
\centering
\includegraphics[page=1,width=\textwidth]{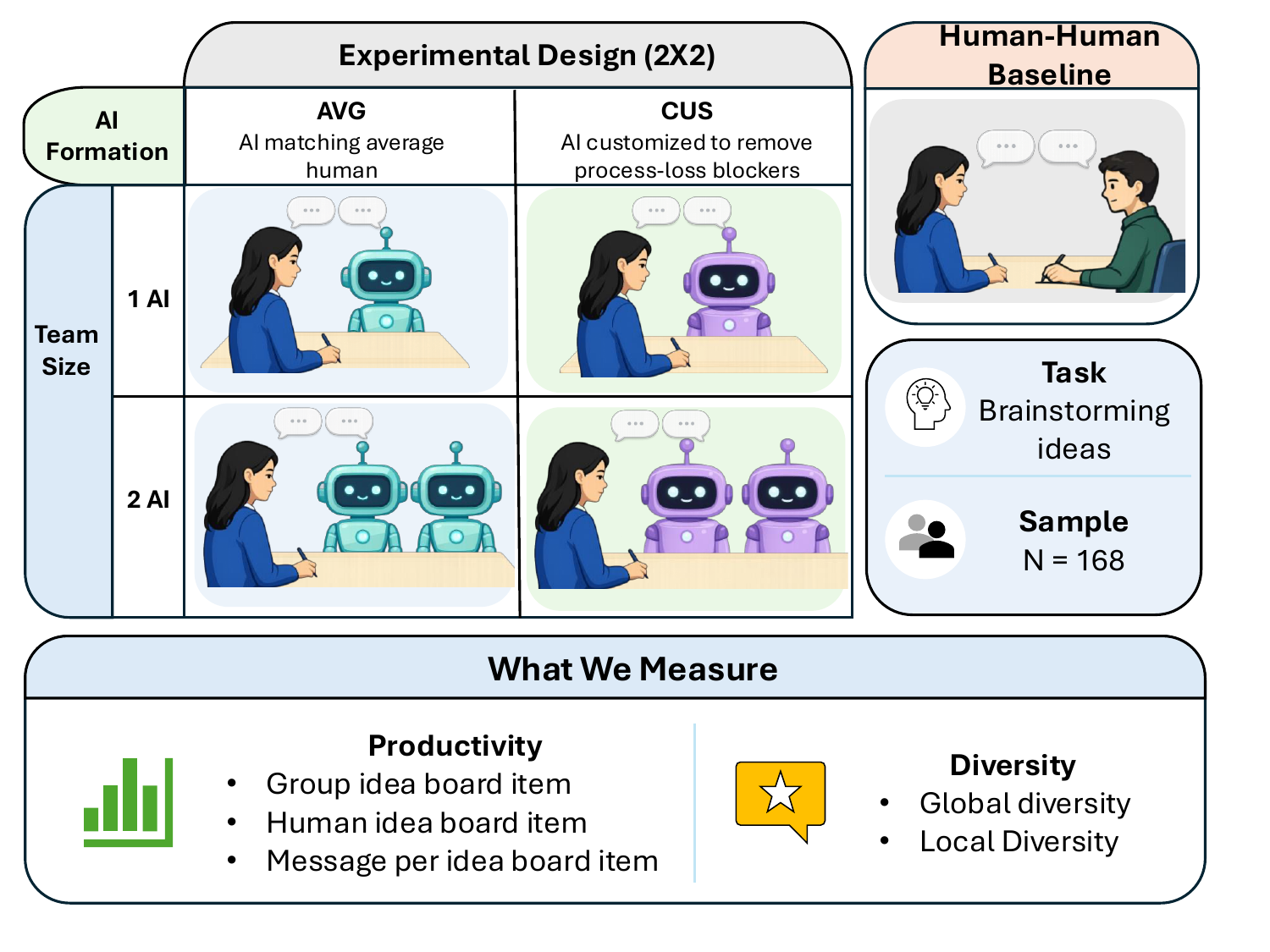}
\Description{Illustration of the study design. The experiment crosses team size, one or two AI partners, with AI formation, average-calibrated or customized, alongside a human-human baseline.}
\caption{Overview of the experimental design. The study crosses AI team size (one versus two partners) with AI formation (AVG versus CUS), alongside a human-human reference condition.}
\label{fig:teaser-design}
\end{teaserfigure}

\maketitle

\section{Introduction}

\begin{figure}[t]
\centering
\includegraphics[width=\linewidth]{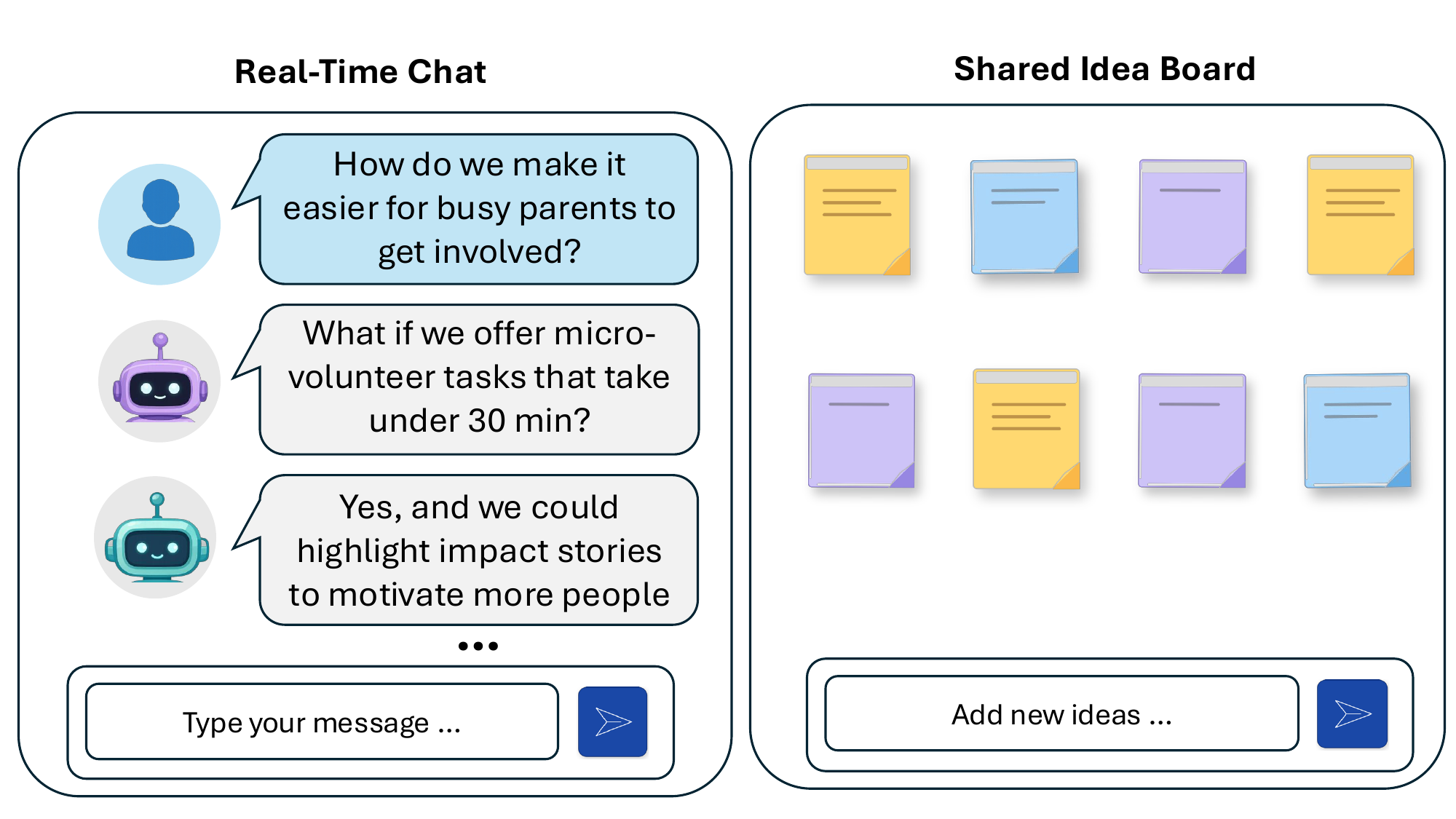}
\Description{Illustration of the brainstorming interface with real-time chat on the left and a shared idea board on the right. Human and AI messages appear in chat while board items accumulate as shared artifacts.}
\caption{Collaborative brainstorming workspace. Human participants and AI partners exchanged messages in real time while contributing ideas to a persistent shared board.}
\label{fig:intro-workspace}
\end{figure}

Generative AI has rapidly improved individual knowledge-worker productivity across writing, customer service, and consulting tasks \cite{noyZhang2023Productivity, brynjolfssonLiRaymond2023Generative,dellAcquaEtAl2026Jagged}. Individual productivity gains, however, do not transfer directly to collaborative settings. When AI is designed to join a team, groups' interactional processes and thus the group's total and per-human outcomes can also shift. Some initial observations suggested that while human-AI teams can potentially lead to more output per worker, these outputs can be more similar, reducing the diversity and creativity of outcomes \cite{juAral2026Collaborating}. Other work shows that prompting large language models with diverse personas can preserve output diversity, suggesting that whether the ``homogenization effect'' presents may depend on how AI perspectives are configured rather than on AI use alone \cite{wanKalman2026DiversePersonas}. Together, these findings raise a set of open research questions: adding AI partners to a team can shift quantity, individual originality, and collective semantic diversity in different directions \cite{ashkinazeEtAl2025AIIdeas, wangEtAl2011Diversity}, yet it remains unclear how group size and AI behavioral configuration jointly shape these outcomes. These tensions are especially visible in creativity-oriented group knowledge work, such as group brainstorming, where AI-infused human-AI brainstorming may affect different parts of group outcome, from idea generation, to idea selection and evaluation given complex cognitive operations (e.g., recall and reasoning in ideation) and social perceptions and motivations (e.g., peer evaluation and social loafing) involved \cite{mullerEtAl2024GroupBrainstorming,shaerEtAl2024Brainwriting}.

Part of what renders human-AI brainstorming distinct and essential to the studies of human-AI interactions is that it depends not only on whether AI is present, but on how AI behaves, and how it affects people individually and collectively. For example, participants may find the ideas an AI agent generated helpful for sustaining the thinking flow, yet they could simultaneously have concerns about feeling blocked and interrupted by the proactiveness or even aggressiveness of the agent, leading to social frustrations \cite{houdeEtAl2025Controlling}. This suggests that an AI agent's behavioral profile, not just its presence or participation, could determine how human team members benefit from agent participation at each stage of group work. If behavioral style matters with a single AI, it could have a larger and more sophisticated consequence when a team includes multiple AI partners.

Adding more AI partners to a brainstorming team as a design act can potentially affect the group dynamics of who thinks what and who does what. Additional agents can generate more candidate ideas, write more items to the shared workspace, and may potentially reach the group's target faster by having more ``headcounts'', but they can also crowd out human contributions, shift the human into a different role, or transform the team dynamics toward a better or worse state. Whether these shifts help or hurt can depend in part on how the team handles three common social ``productivity blockers'' or social psychological mechanisms that shape group brainstorming processes and outcomes. \emph{Competitive orientation} leads a group member to prioritize advocating for their own ideas rather than building on what others contribute \cite{deutsch1949}. \emph{Evaluation apprehension} causes members to withhold contributions for fear of negative judgment from the group \cite{paulus1993, bolinNeuman2006Personality}. \emph{Social loafing} leads individuals to reduce effort when their contributions are less identifiable within the collective output \cite{paulus1993, mulveyKlein1998Loafing}. These blockers are well-documented in human groups, but whether tuning an AI partner's behavior along these dimensions might alleviate or amplify them remains untested. Several recent studies have explored multi-agent AI configurations for creative work \cite{quanEtAl2026Colleagues, limEtAl2026TeamFormation, songEtAl2025MultiAgents,ghoshRintel2025YesAnd,nomuraEtAl2024CollaborativeBrainstorming}, but existing work did not isolate the factor of group size from multi-agent formation, and also did not examine the role of social blocking mechanisms. Similarly, in human-AI collaboration, whether an AI partner should be calibrated to match humans' brainstorming behaviors and dynamics, and how such calibration would influence group brainstorming comparable to human-only groups also remains unknown.

Taken together, these gaps leave certain key research questions unanswered. When does adding a plurality of AI partners actually help? Do the benefits simply scale with the number of agents, depend on how each agent is configured (whether calibrated to match average human brainstorming behavior or customizable by human workers on these social blocking dimensions), or on the interaction between the two? To address this, we recruited 168 participants into a randomized 2x2 study that crosses group size (one versus two AI partners) with AI formation (AVG versus CUS), alongside a same-study human-human reference. In the AVG condition, each AI partner is calibrated to match average human brainstorming behavior. In the CUS condition, participants configure their AI partners' behavioral profile on the three blocker dimensions before the session begins; in two-AI sessions, both agents adopt the same configuration. The group size factor tests whether adding a second AI partner further changes group output and the human contribution pipeline. The formation factor tests whether giving the human control over the AI's behavioral profile changes how group size affects the team.

Within this design, we ask two questions:

\textbf{RQ1:} Does adding more AI partners help productivity?

\textbf{RQ2:} Does adding more AI partners increase the quality of outcome?

For productivity, one AI partner produces a total group idea count on the board comparable to the human-human reference, but adding a second AI raises the total while reducing the number of ideas the human adds to the board. With only one AI partner, formation matters: humans in the AVG condition add more ideas to the board than in any other condition, including the human-human reference, while humans in the CUS condition add fewer. With two AI partners, this formation difference disappears and both converge near the same lower level of human-added board items. The number of human messages per human board item rises with a second AI, yet human chat rates and novel-message rates show no significant decline, suggesting that the human is not disengaging but that fewer of their ideas are making it onto the board. The two formations also operate in different interaction modes: AVG AI partners co-generate alongside the human with substantial social and filler talk, while CUS AI partners adopt a coaching style with more evaluative and questioning messages. Across all AI conditions, humans shift away from social conversation toward generative output compared to human-human groups.

For output quality, both one-AI and two-AI groups produce human-authored board items that are more semantically diverse than those in human-human groups across both global and local measures. For the global measure, which compares human-authored items against the full study-wide idea pool, the diversity gain appears with any AI but does not increase further with a second AI. For the local measure, which compares human-authored items against their own room's board ecology, adding a second AI further increases differentiation. This pattern suggests that AI reshapes the room-level ecology around the human's contributions rather than broadening what the human produces across the study as a whole.

These findings carry three implications for the design of AI-augmented teams. First, the impact of group size is not uniform: productivity patterns that hold with one AI partner do not transfer reliably to two, so designers should treat each team configuration as a distinct design case rather than extrapolating from single-agent results. Second, adding more AI partners shifts who writes ideas to the board without reducing how much the human talks, which means future systems aiming to preserve human authorship will need mechanisms beyond encouraging participation. Third, while any AI presence increases the global diversity of human-authored output, only adding a second AI increases local differentiation, suggesting that future work should examine how multi-agent configurations can be designed to raise both global and local quality simultaneously rather than trading one for the other.

\section{Related Work}

Adding AI partners to creative group work can change who contributes, what gets recorded, and how the resulting ideas are evaluated. At the individual level, AI assistance has improved speed and output quality \cite{noyZhang2023Productivity, brynjolfssonLiRaymond2023Generative}, but these gains are contingent. Consultants who used AI on tasks outside its capabilities performed worse than those without \cite{dellAcquaEtAl2026Jagged}, and LLM assistance that boosted creativity during assisted tasks hindered independent creative performance afterward \cite{kumarEtAl2025HumanCreativity}. At the team level, human-AI dyads in a field experiment produced more per worker while also producing more homogeneous outputs \cite{juAral2026Collaborating}. Creative outcomes sharpen this divergence, as collective idea diversity can increase under AI exposure without a corresponding gain in individual originality \cite{ashkinazeEtAl2025AIIdeas}. Whether this divergence occurs may depend on how the AI partner is configured, as diversifying AI personas in a story-writing experiment preserved collective diversity relative to a human-only baseline \cite{wanKalman2026DiversePersonas}. This distinction matters for group brainstorming because quantity and semantic breadth are only moderately correlated \cite{wangEtAl2011Diversity}. Any study of AI in creative teams therefore needs to separate these dimensions rather than collapsing them into one productivity number.

Existing studies have not disentangled the effect of adding more AI partners from the effect of changing how each AI partner behaves, leaving open whether the number of partners, their behavioral formation, or the combination drives the observed shifts in human contribution. This problem predates generative AI. Research on shared digital workspaces showed that awareness of co-workers' activity shapes how people coordinate contributions \cite{gutwinGreenberg2002Workspace}. With even one AI partner candidate generation and selection are already different stages. An AI partner can supply most ideas produced while humans select far fewer of them as final \cite{mullerEtAl2024GroupBrainstorming}, and proactive AI behavior can sustain conversation flow while simultaneously overwhelming participants \cite{houdeEtAl2025Controlling}. Studies of human-AI co-creation similarly identify turn-taking, initiative, and contribution type as variables that shape perceived agency and idea selection \cite{rezwanaMaher2023COFI,qinEtAl2025Timing,youngNgEtAl2025IdeaSelection}, yet in each case the AI's behavioral profile is varied without independently manipulating how many AI partners are present. Multi-agent systems extend these dynamics but bundle agent count with changes in personas, roles, workflow, and interface \cite{quanEtAl2026Colleagues, limEtAl2026TeamFormation}. Adjacent examples include multi-agent ideation systems that simulate diverse perspectives, use confidence-based turn-taking, or structure brainstorming through issue/idea/pro/con agent roles, as well as interface designs that make an AI participant socially prominent in group discussion \cite{ghoshRintel2025YesAnd,nomuraEtAl2024CollaborativeBrainstorming,johnsonEtAl2026SocialProminence}. We address both gaps by crossing group size (one versus two AI partners) with formation (calibrated to average human behavior versus configured by the participant), while tracking conversation, board authorship, and semantic differentiation as separate outcomes.

When both humans and AI partners can write to the same board, and every item counts toward the same group target, who records an idea becomes a consequential design variable. A human may propose many ideas in conversation while only a fraction become board items, and those board items may be written by the human, by the AI partner, or prompted by one actor and recorded by the other. The board also shapes how quality is measured. Because AI-written items both help complete the target and change the comparison set for later human items, a human contribution can become more semantically distinct within its own room without becoming more distinct from the study-wide pool \cite{wangEtAl2011Diversity, ashkinazeEtAl2025AIIdeas}. The shared-artifact setting also makes control a design variable, as prior automation work distinguishes systems that gather information, select actions, and implement actions, a useful lens for separating AI suggestions from AI-authored board items \cite{parasuramanEtAl2000Automation}. This leaves open whether classic group brainstorming losses, such as evaluation apprehension and social loafing \cite{paulus1993}, still operate when the partner is an AI agent rather than a human, or whether configuring the AI on these dimensions changes the outcome. To our knowledge, no prior study isolates AI headcount from agent behavioral formation in a shared-artifact brainstorming task, which is what makes it possible to ask whether adding a second AI partner changes group output, human board authorship, and idea quality independently of how each AI partner is configured.

\section{Method}

\subsection{Design and Participants}

We conducted a 2 \(\times\) 2 between-subjects experiment crossing group size (one versus two AI partners) with AI formation (AVG versus CUS), alongside a same-study human-human reference. The group size factor tests whether adding a second comparable AI changes group output and the human contribution pipeline. The formation factor tests whether giving the human control over the AI's behavioral profile changes how group size affects the team.

We recruited 168 participants through Cloud Research and compensated each \$8 for approximately 45 minutes. Of these, 120 were randomly assigned to one of four conditions in the 2 \(\times\) 2 factorial (30 per cell), each brainstorming individually with one or two AI partners in a single session. An additional 48 participants completed the same task in 24 human-human dyads, providing a same-study human-human reference point outside the factorial design. Across all conditions, 46.9\% identified as male, 52.5\% as female, and 0.6\% as non-binary or preferred not to say; 96.6\% reported native or near-native English proficiency. Each participant completed two brainstorming rounds followed by a post-task survey. The study was approved by the Institutional Review Board of the University of California, Davis.

\subsection{Task and Platform}

Each session consisted of two brainstorming rounds, following the open-ended practical-problem format used in classic brainstorming research where topics require no specialized knowledge \cite{osborn1957, paulus1993, paulus2007}. In the first round, groups generated ideas for how a local school district could increase volunteer engagement from its community. In the second round, groups generated ideas for improving online learning in pre-university summer courses for high school students earning university credit.

Groups worked through a custom web platform that presented a shared workspace with a scrolling chat panel and a persistent idea board visible to all members (Figure \ref{fig:intro-workspace}). Participants typed messages in the chat and saw their own and their partners' contributions appear in chronological order with display names and timestamps. Any group member could post a discrete idea to the board at any time; items remained visible and could be referenced throughout the session. Both human participants and AI partners could contribute chat messages and board items, and all content appeared in real time. A running counter tracked the group's progress toward a 40-item target, and an optional button to advance to the next round appeared once the threshold was reached; participants could continue contributing before choosing to proceed.

\subsection{Procedure}

After providing informed consent and reading a brief task introduction, participants in the CUS conditions completed a structured walkthrough with their AI partner before brainstorming began (described in Section~\ref{sec:formations}); participants in the AVG conditions and human-human dyads proceeded directly to the brainstorming workspace. All participants then completed two consecutive brainstorming rounds on the two topics, each beginning with a topic prompt displayed at the top of the workspace and a greeting message from the AI partner (or partners); human-human dyads received only the topic prompt. Sessions lasted approximately 45 minutes on average, including the CUS walkthrough where applicable. The session concluded when both rounds were complete, after which participants completed a post-task survey administered through Qualtrics.

\subsection{AVG and CUS Formations}
\label{sec:formations}
The two formations represent contrasting design strategies for AI brainstorming partners. In the AVG formation, each AI is calibrated to match average human brainstorming behavior, providing a fixed behavioral baseline against which to evaluate participant-directed customization. In the CUS formation, each participant configures the AI's behavioral profile on these three dimensions before the session begins, setting how assertively the AI advocates for its own ideas versus building on others' (competitive orientation), how readily it shares contributions without self-censoring (evaluation apprehension), and how consistently it contributes substantive content rather than minimal responses (social loafing). Crossing these two formations with group size makes it possible to ask whether the effect of adding a second AI depends on how each agent is configured.

\textbf{AVG formation.} The AVG condition approximated a typical human brainstorming partner. Each AVG agent was calibrated against a 2,697-message human-human corpus from the same brainstorming task: the agent's prompts targeted the observed distribution of message types (67\% elaboration, 11\% social, 11\% question, 5\% backchannel), message length variance, and consecutive-message patterns. For board contributions, AVG agents drew candidate items from a pre-populated idea bank of human-generated ideas. The observed behavioral profiles of both formations are reported in Section~\ref{sec:ai-profiles}.

\textbf{CUS formation.} The CUS condition gave participants direct control over the AI's behavioral profile on the three process-loss dimensions described in the introduction. Before the brainstorming session, each participant completed a structured walkthrough in which the AI guided them through nine phases: an introduction, followed by level selection (Low, Medium, or High) and a free-text description for each of the three dimensions, then a summary and confirmation step.

The three dimensions mapped onto the process-loss blockers as follows. \emph{Competitive orientation} governed how assertively the AI advocated for its own ideas versus building on the human's contributions: Low prioritized the partner's ideas, Medium balanced both parties' contributions, and High led the AI to actively champion its own directions. \emph{Evaluation apprehension} governed how unconventional the AI's suggestions could be: Low restricted output to practical, established approaches; Medium introduced creative ideas framed with hedging qualifiers; High permitted the AI to propose ideas freely regardless of how unconventional they appeared. \emph{Social loafing} governed how much substance the AI contributed per turn: Low produced brief acknowledgments and selective contributions, Medium usually yielded substantive responses, and High produced substantive new content on every turn.

After the walkthrough, the participant's chosen levels and free-text descriptions were compiled into the AI's behavioral instructions, which guided the agent's behavior throughout the session. In two-AI CUS sessions, both agents adopted the same configuration set by the human participant.

\subsection{AI Implementation}

All AI partners were powered by OpenAI's GPT-4o. Each agent received the full conversation history on every turn and produced a single structured response containing both a chat message and, when appropriate, a board item. In two-AI sessions, the system enforced a turn policy in which each agent contributed at most one message per human turn, consistent with the coordination strategies adopted in recent multi-agent brainstorming studies \cite{quanEtAl2026Colleagues, limEtAl2026TeamFormation}. In two-AI sessions, agents could not distinguish which messages came from the human participant and which came from the other AI; both appeared identically in the conversation feed.

\subsection{Measures}

\label{sec:measures}

\textbf{Group board items.} The total number of board items per session at session end, including both human-authored and AI-authored contributions. Each session contributed one count.

\textbf{Human board items.} The number of human-authored board items per person within each session. For human-human dyads, the session total was divided by two. Each session contributed one value.

\textbf{Human messages per board item.} The ratio of human chat messages to human-authored board items within a session. Higher values indicate that the human engaged in more conversational activity per visible contribution to the board.

\textbf{Message-act coding.} We classified every chat message from both AI and human participants into one of four act categories developed for this study. The categories reflect the principle that idea generation and idea evaluation are distinct brainstorming phases that benefit from separation \cite{paulus2007}, as well as the observation that AI brainstorming agents can take on generative, evaluative, or social roles within a single session \cite{mullerEtAl2024GroupBrainstorming}. \emph{Generative} messages proposed new ideas, elaborated on existing ones, or offered substantive extensions to a prior contribution. \emph{Evaluative} messages assessed the merit of a contribution, critiqued an approach, or endorsed another participant's idea. \emph{Questioning} messages asked the other participant for input, sought clarification on an idea, or solicited direction for the session. \emph{Social and filler} messages consisted of greetings, backchannels, encouragement, and off-task talk that did not advance the brainstorming content. Classification followed a deterministic rule-based scheme that applied message length thresholds, punctuation cues, and keyword matching.

\textbf{Global and Local Diversity.} For each human-authored board item, we computed \(\log(1 / \bar{s})\), where \(\bar{s}\) is the mean pairwise cosine similarity between the focal item's sentence embedding (all-MiniLM-L6-v2, 384 dimensions) and all other board items in a reference pool, following the breadth-of-concepts measure introduced by Wang, Fussell, and Cosley \cite{wangEtAl2011Diversity}. We report two variants. \emph{Global diversity} compares each item against the full board-item pool across all 120 AI-condition sessions. \emph{Local diversity} compares each item against only the board items within its own session. Both variants were computed at the room level by averaging across all human-authored items within each session.

\section{Results}

\subsection{Does Adding More AI Help Productivity? (RQ1)}

Groups with two AI partners produced more board items overall but fewer human-authored items, and each human item was accompanied by more conversation. Formation mattered with one AI partner but ceased to matter with two. The human-human reference dyads provided a same-study baseline for the four AI conditions.

\subsubsection{Total group board items}

Two-AI rooms accumulated more board items than one-AI rooms (55.8 vs 41.8, \(d\)=+1.25, \textbf{\(p\)\textless.0001***}). One-AI rooms did not significantly differ from the human-human reference (41.8 vs 40.9, \(d\)=+0.37, \(p\)=.064), indicating that a single AI did not inflate group output beyond what two humans produced together.

\subsubsection{Human board items}

Although groups with more AI partners produced more items overall, the human's own contribution to the board was substantially lower with two AI partners than with one (13.5 vs 21.7 items per person, \(d\)=-0.95, \textbf{\(p\)\textless.0001***}). Formation moderated this pattern when only one AI was present: AVG participants wrote more items than any other condition, exceeding even the human-human reference (25.7 vs 20.5, \(d\)=+1.60, \textbf{\(p\)\textless.0001***}), while CUS participants fell below it (17.7 vs 20.5, \(d\)=-0.62, \textbf{\(p\)=.018*}). With two AIs, both formations converged near 13 items and the formation difference vanished (13.7 vs 13.3, \(d\)=-0.05, \(p\)=.862). The factorial estimands make this moderation explicit (Table \ref{tab:human-board-factorial}).

\begin{longtable}[]{@{}lrrr@{}}
\caption{Factorial estimands for human-authored board items per person, using OLS estimates (\(N\)=120). \label{tab:human-board-factorial}}\tabularnewline
\toprule\noalign{}
Term & \(\beta\) & 95\% CI & Holm \(p\) \\
\midrule\noalign{}
\endfirsthead
\toprule\noalign{}
Term & \(\beta\) & 95\% CI & Holm \(p\) \\
\midrule\noalign{}
\endhead
\bottomrule\noalign{}
\endlastfoot
Two-AI marginal & -8.20 & {[}-11.21, -5.19{]} & \textbf{\textless.001***} \\
Customization marginal & -4.23 & {[}-7.25, -1.22{]} & \textbf{.025*} \\
Count \(\times\) formation & +7.53 & {[}1.50, 13.56{]} & \textbf{.044*} \\
\end{longtable}

\subsubsection{Human messages per human board item}

The lower human board-item count in two-AI rooms did not reflect disengagement from the conversation. Participants in two-AI rooms continued to talk at comparable rates, but each human-authored board item was accompanied by more chat: 5.42 messages per board item compared with 2.52 in one-AI rooms (\(d\)=+0.88, \textbf{\(p\)\textless.0001***}). AVG participants with one AI converted chat into board items more efficiently than even the human-human reference (1.91 vs 2.45 messages per item, \(d\)=-0.78, \textbf{\(p\)=.006**}). Participants paired with one CUS AI required more conversation per item than those with one AVG AI (3.12 vs 1.91, \(d\)=+0.94, \textbf{\(p\)=.001***}), whereas in two-AI groups both formations averaged five to six messages per item. This shifting ratio of chat to board items prompted us to model the specific AI behaviors predicting human authorship.

\begin{longtable}[]{@{}lrrrr@{}}
\caption{Human messages per human board item: pairwise contrasts across conditions.}
\label{tab:messages-per-item}\\
\toprule\noalign{}
Contrast & Means & Difference & \(d\) & \(p\) \\
\midrule\noalign{}
\endhead
\bottomrule\noalign{}
\endlastfoot
2AI vs 1AI & 5.42 vs 2.52 & +2.90 & +0.88 & \textbf{\textless.0001***} \\
1CUS vs 1AVG & 3.12 vs 1.91 & +1.21 & +0.94 & \textbf{.001***} \\
2CUS vs 2AVG & 6.08 vs 4.76 & +1.32 & +0.30 & .251 \\
1AVG vs HH & 1.91 vs 2.45 & -0.54 & -0.78 & \textbf{.006**} \\
\end{longtable}

\subsubsection{Predictors of human board authorship}

AI board volume predicted fewer human board items, and the share of social and filler AI talk predicted more. We coded each AI message into one of four act categories: \emph{generative} (proposing or elaborating on ideas), \emph{evaluative} (assessing or critiquing contributions), \emph{questioning} (asking the human for input or clarification), and \emph{social/filler} (greetings, backchannels, and off-task talk). We then ran separate OLS models, each entering one category's share alongside AI board volume, group size, and formation as controls (\(N\)=120). Across all four models, each additional AI-authored board item was associated with approximately 0.27 fewer human board items (\textbf{\(p\)\textless.001***}). Of the four act categories, only the share of social and filler talk reached significance (unstandardized \(\beta\)=+26.11, \textbf{\(p\)=.029*}): rooms where the AI devoted a larger share of talk to low-pressure social conversation had more human board items. The shares of generative, evaluative, and questioning messages were individually non-significant (full coefficient tables in Appendix~\ref{app:ols-models}). These varying conversational correlates naturally arise from the distinct behavioral profiles of the AI formations, which we examine next.

\begin{figure}
\centering
\includegraphics[width=\linewidth]{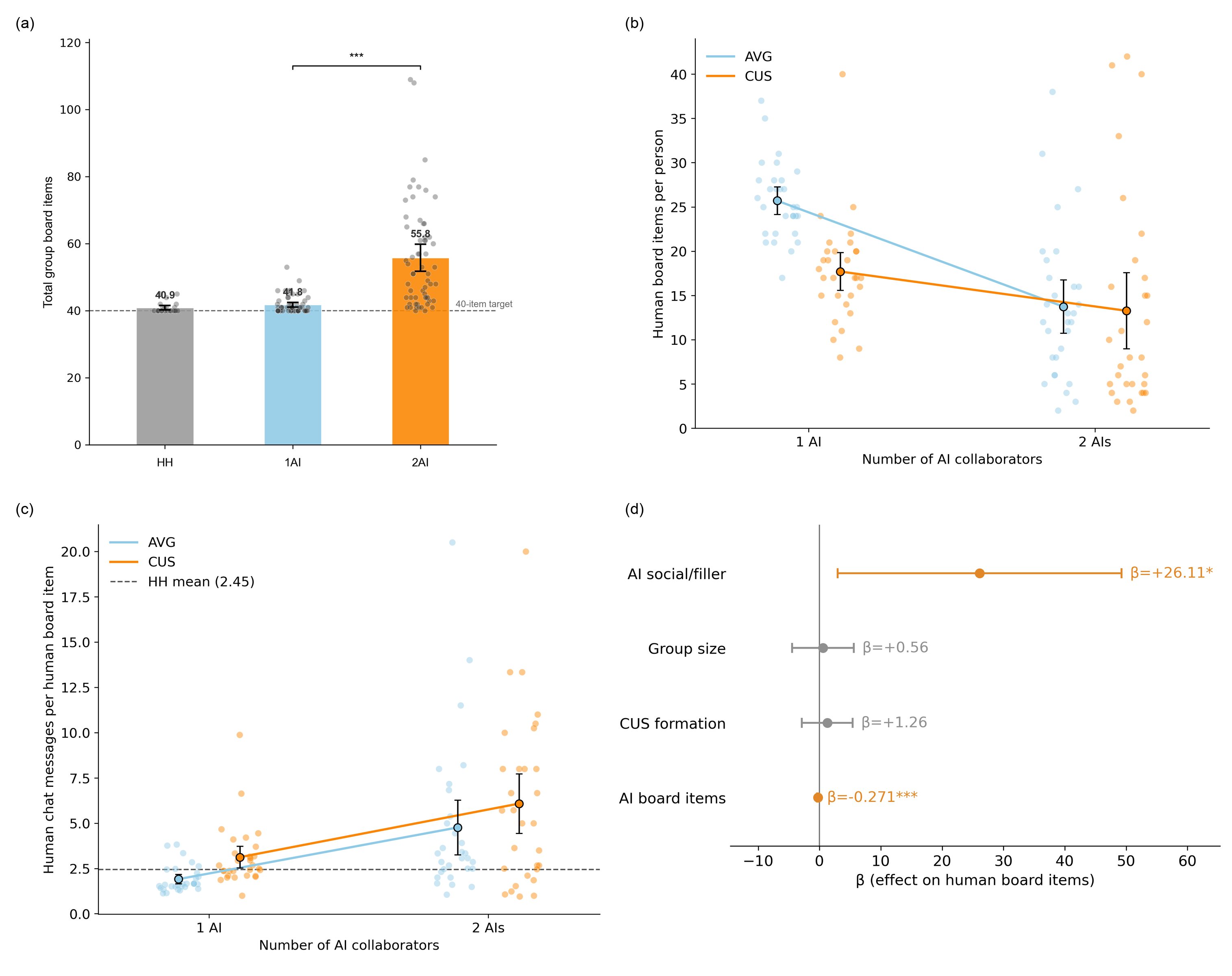}
\Description{Four-panel results figure showing group board volume, terminal human board items, human messages per board item, and the social/filler model forest plot.}
\caption{Productivity and human-authorship results. Panel (a) shows group board items by condition with the 40-item quota. Panel (b) shows human-authored board items per person. Panel (c) shows human messages per human board item. Panel (d) shows the OLS model linking AI board volume and social/filler talk to human board items.}
\label{fig:rq1-productivity-panel}
\end{figure}

\subsubsection{AI behavioral profiles and human composition}
\label{sec:ai-profiles}

The two formations produced distinct interaction modes. With one AI partner, the AVG agent adopted what we term a \emph{conversational co-generator} profile: 58\% of its messages proposed or elaborated on ideas, and 22\% were social or filler talk, closely matching the human-human generative share while retaining more social/filler talk. The CUS agent, by contrast, operated as a \emph{coaching} profile: 25\% of its messages evaluated ideas and 32\% posed questions, with less than 1\% social filler. With two AI partners, CUS agents maintained their coaching profile (23\% evaluative, 33\% questioning), but AVG agents shifted substantially: generative messages rose from 58\% to 77\% while social filler dropped from 22\% to 1.4\%. The two formations still differed on every message-act category at both group sizes (all CUS-vs-AVG contrasts \(p \leq .002\)), yet the qualitative character of the AVG profile changed. With one AI partner, AVG balanced idea generation with social conversation; with two, it concentrated almost entirely on generation, narrowing the gap with CUS on social talk from a 21-percentage-point difference to just over one. CUS AI partners also authored more board items than AVG AI partners at both group sizes (all \textbf{\(p \leq .002\)**}; Table \ref{tab:ai-profile-contrasts}).

\begin{table}[t]
\centering
\small
\caption{CUS-versus-AVG AI behavioral profile contrasts by AI count.}
\label{tab:ai-profile-contrasts}
\begin{tabular}{@{}lrrr@{}}
\toprule
Metric & AVG & CUS & CUS--AVG \(d\), \(p\) \\
\midrule
\multicolumn{4}{@{}l}{\textbf{One AI partner}} \\
AI board items & 16.5 & 23.7 & \(d\)=+1.41, \textbf{\(p\)\textless.001***} \\
AI generative \% & 58.1\% & 42.7\% & \(d\)=-1.27, \textbf{\(p\)\textless.001***} \\
AI evaluative \% & 2.8\% & 25.0\% & \(d\)=+2.65, \textbf{\(p\)\textless.001***} \\
AI questioning \% & 17.2\% & 31.7\% & \(d\)=+0.97, \textbf{\(p\)\textless.001***} \\
AI social/filler \% & 21.9\% & 0.7\% & \(d\)=-4.33, \textbf{\(p\)\textless.001***} \\
\midrule
\multicolumn{4}{@{}l}{\textbf{Two AI partners}} \\
AI board items & 36.4 & 48.2 & \(d\)=+0.89, \textbf{\(p\)=.002**} \\
AI generative \% & 77.1\% & 44.1\% & \(d\)=-3.75, \textbf{\(p\)\textless.001***} \\
AI evaluative \% & 6.2\% & 23.2\% & \(d\)=+3.16, \textbf{\(p\)\textless.001***} \\
AI questioning \% & 15.4\% & 32.5\% & \(d\)=+1.43, \textbf{\(p\)\textless.001***} \\
AI social/filler \% & 1.4\% & 0.2\% & \(d\)=-1.09, \textbf{\(p\)\textless.001***} \\
\bottomrule
\end{tabular}
\end{table}

The presence of an AI partner fundamentally altered how humans communicated. Social and filler talk nearly disappeared, dropping from 15\% of messages in human-human groups to 1 to 3\% across all four AI conditions. Where that freed capacity went depended on team size: in two-AI rooms, humans redirected it toward idea generation (71\% and 69\% generative vs 58\% in human-human groups), while in one-AI rooms they directed it toward questioning the AI partner (19\% in 1AVG and 14\% in 1CUS vs 7\% in human-human groups). This compositional shift, together with the stable chat rates and lower board authorship reported above, is consistent with a pattern in which AI partners changed what the human thought and did within the group rather than how much the human participated.

\begin{figure}
\centering
\includegraphics[width=\linewidth,height=.6\textheight,keepaspectratio]{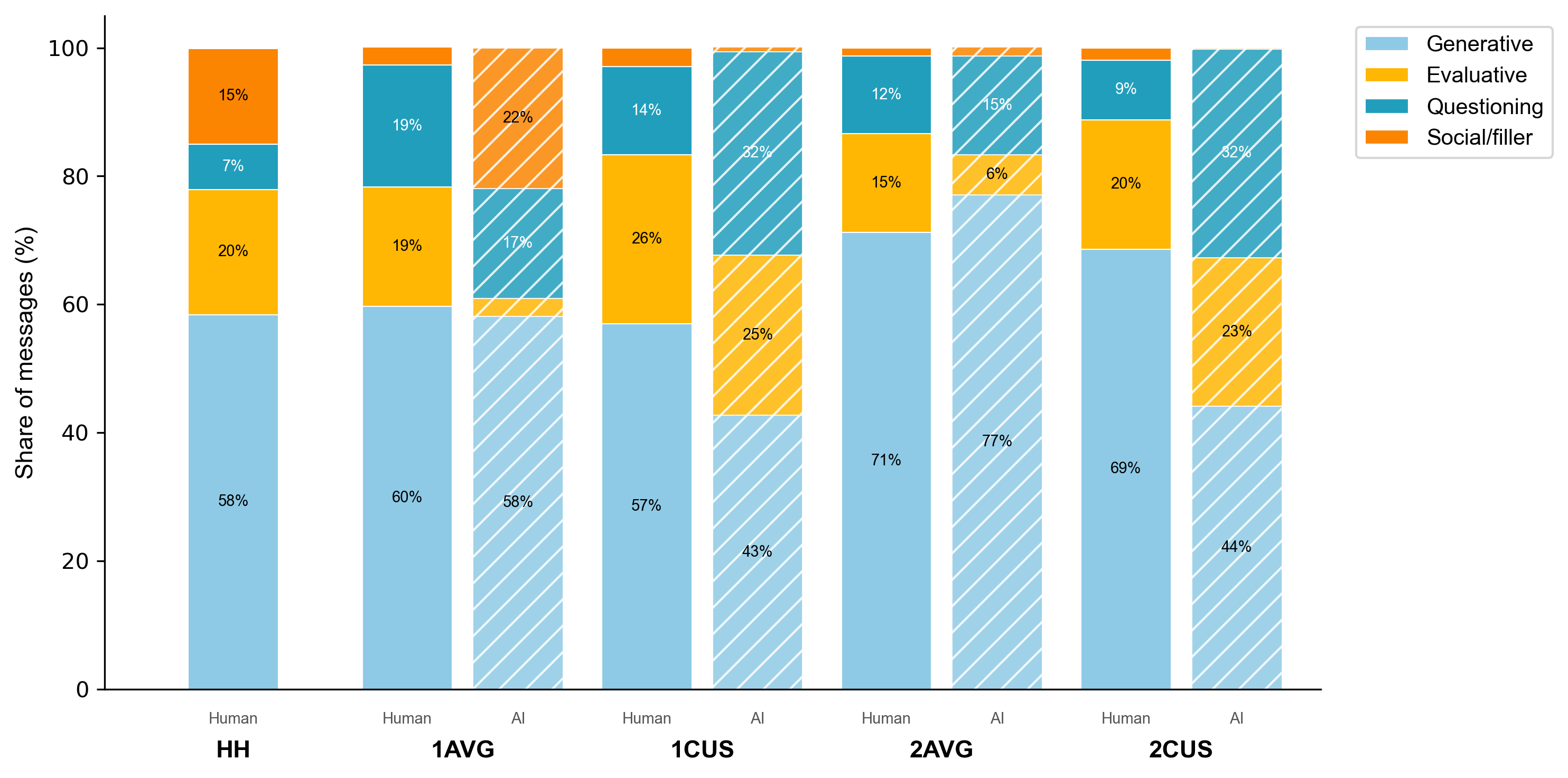}
\Description{Paired human-vs-AI composition}
\caption{Paired human and AI message-act composition by condition. Stacked bars show the share of generative, evaluative, questioning, and social/filler messages. HH has only a human bar.}
\end{figure}

Humans and AI partners occupied complementary rather than overlapping communicative roles, and the nature of that complementarity depended on formation. In 1AVG rooms, human and AI generative shares were nearly identical (60\% vs 58\%), but the two diverged on their secondary function: the AI devoted 22\% of messages to social talk while the human devoted 19\% to questions. In CUS rooms the separation was sharper, with CUS AIs concentrating on questioning (32\%) and humans contributing more generative content (57 to 69\%).

\subsection{Does Adding More AI Increase Output Quality? (RQ2)}

Human-authored board items in all AI conditions were more semantically diverse than those in human-human groups, but rooms with two AIs showed further gains only when diversity was measured locally (against the session's own board) rather than globally across the study.

\subsubsection{Global diversity}

Human-authored items in all four AI conditions were more globally diverse than those in human-human groups (\(d\)=+0.57 to +1.08, all \textbf{\(p\)\textless.05*}), but two-AI rooms were not more globally diverse than one-AI rooms (0.839 vs 0.832, \(d\)=+0.25, \(p\)=.182). Global diversity compared each human-authored board item's sentence embedding against the full board-item pool across all 120 AI-condition sessions (Section \ref{sec:measures}). Any AI presence broadened the semantic range of human contributions relative to an all-human group, yet a second AI did not extend that range further.

\subsubsection{Local diversity}

Where global diversity showed no two-AI gain, local diversity did. Two-AI rooms produced locally more diverse human items than one-AI rooms (0.824 vs 0.802, \(d\)=+0.57, \textbf{\(p\)=.002**}), where local diversity compared each human-authored item against only the other board items within its own session. All AI conditions exceeded the human-human reference (\(d\)=+0.61 to +1.28, all \textbf{\(p\)\textless.05*}). Formation did not explain local diversity within a fixed group size (1AVG vs 1CUS: \(d\)=-0.36, \(p\)=.164; 2AVG vs 2CUS: \(d\)=-0.15, \(p\)=.575).

\begin{figure}
\centering
\includegraphics[width=\linewidth,height=.6\textheight,keepaspectratio]{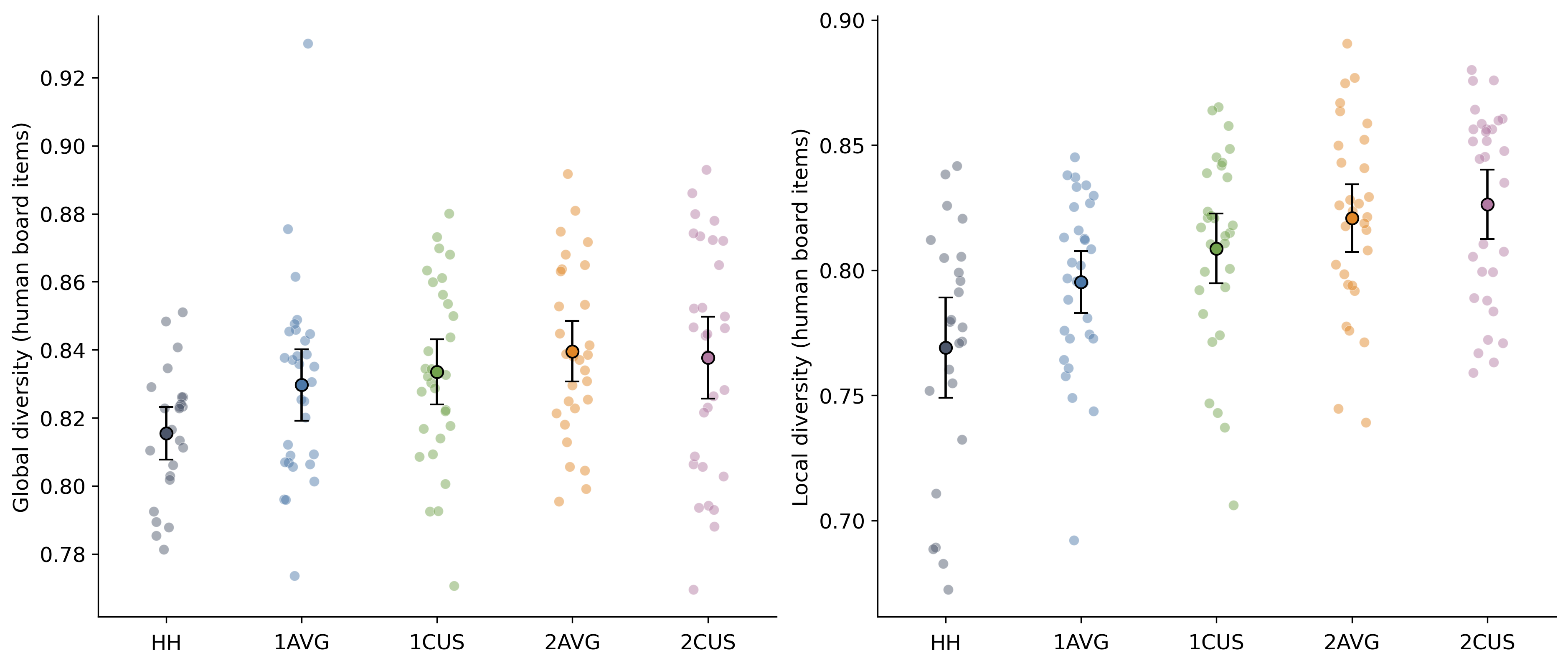}
\Description{Global and local diversity}
\caption{Global and local differentiation of human-authored focal board items. Error bars show 95\% confidence intervals. The two-AI versus one-AI difference is detected locally but not globally.}
\end{figure}

\section{Post-Hoc Analyses and Discussions}

As AI partners are increasingly designed to participate in creativity-oriented group knowledge work, understanding how they change what humans think and do within a team is as important as whether they improve what the team produces. Yet most evidence for AI-driven productivity comes from individual tasks or single-agent pairings, leaving open whether those gains hold when a plurality of AI partners is designed to join a group and whether it matters if each AI partner mimics typical human brainstorming behavior or is customized by the participant to reduce known social "productivity blockers". By crossing the number of AI partners (one versus two) with their behavioral formation (calibrated to average human behavior versus customized by the participant) in a controlled brainstorming experiment, we ask two questions. Does adding a plurality of AI partners help group outcomes (RQ1)? Does adding more AI partners increase the quality of outcomes (RQ2)?

\subsection{Group output gains do not transfer to human authorship}

\textbf{Group output.} Adding more AI partners can increase group productivity. Two-AI rooms produced more distinct items on the shared board than one-AI rooms (55.8 vs 41.8, \(d\)=+1.25), and one-AI rooms already matched the human-human reference (41.8 vs 40.9, \(d\)=+0.37, \(p\)=.064).
Prior multi-agent studies have reported similar output gains, with one system finding higher perceived quality and novelty when multiple AI personas collaborated on ideation \cite{quanEtAl2026Colleagues}.
The scale of this output gain varies with the collaborative setting. In a large field experiment on marketing ad creation, teams with a single AI partner produced 50\% more output per worker than human-only teams \cite{juAral2026Collaborating}, whereas in our brainstorming setting, one AI partner matched one human in total board output (\(d\)=+0.37, \(p\)=.064), and the substantial group-level increase required adding a second AI partner. Despite this difference in output magnitude, both studies converge on a common consequence: as AI partners are added, humans continue communicating at comparable rates while the AI accounts for a growing share of the recorded output. In their ad-creation teams, humans made 62\% fewer direct edits and delegated 17\% more work to the AI; in our brainstorming rooms, the AI's share of board authorship grew from roughly half to three quarters as a second partner was added. The consistency of this delegation pattern across ad creation and real-time brainstorming, two tasks with different AI roles and interaction modalities, suggests that the shift from direct authorship to conversational steering may be a recurring pattern in human-AI collaboration that designers should anticipate rather than treat as a side effect.

This pattern contrasts with a well-established finding in human-only brainstorming, where adding more group members typically does not produce proportional gains because productivity losses increase with group size \cite{diehlStroebe1987, mullenJohnsonSalas1991}. In human-only groups, well-documented productivity losses (production blocking, evaluation apprehension, and social loafing) cause per-person output to decline as the team grows \cite{paulus1993}. In human-AI groups, the dynamic reverses: each additional AI partner generates ideas without incurring these social costs, so group output rises even as the human's share of board authorship falls. This reversal renders the design of human-AI team composition distinct from the design of human-only teams: the interactional process through which this gain is achieved, shifting who thinks what and who does what in the group, reshapes the human's role in ways that group-size research on human-only teams does not predict.

\textbf{Personal authorship.} The flip side of the group-level gain is a decline in what the individual human writes to the board. With two AI partners, human-authored items were substantially lower than with one (13.5 vs 21.7, \(d\)=-0.95), and the group-level gain from a second AI came almost entirely from AI-authored items. As AI-authored items accumulate on the shared board, the marginal return from adding one more human-authored item diminishes. The human does not stop contributing; they shift from direct board authorship toward conversational steering, evaluation, and elaboration. This shift is consistent with what recent work on individual AI-assisted writing terms \emph{cognitive debt}: when users repeatedly rely on AI-generated content, the effortful cognitive processes required for independent production may be replaced rather than supplemented, yielding short-term output gains while deferring the deeper engagement that sustains ownership and retention \cite{kosmynaEtAl2025CognitiveDebt}.

The principle of least collaborative effort explains why a persistent shared board changes the cost of capture. The principle holds that communication requires grounding, the joint process of establishing that what has been contributed is understood, and that participants try to minimize this combined work \cite{clarkBrennan1991Grounding}. How they minimize depends on the medium. In evanescent media such as speech, contributions fade and grounding demands active back-channel confirmation. In a reviewable medium, contributions persist as inspectable artifacts, shifting the cost from confirming understanding to checking a visible record. Because the brainstorming board is highly reviewable, AI-authored items remain visible as part of the group's shared record. The relevant economy is duplicated capture labor rather than literal repetition of the same idea. Conversation becomes the channel where the human can decide what still needs to be added, combined, or refined. This account also makes scale consequential. As AI partners occupy more of the shared artifact, direct human board authorship can decline even while human participation remains visible.

The scale of the authorship shift we observe is moderate compared with what emerges in more agent-rich configurations. In teams with two to five AI agents, humans produced only 1.6\% of all ideas (7 of 451), whereas our participants in two-AI rooms still authored roughly 24\% of board items (13.5 of 55.8) \cite{limEtAl2026TeamFormation}. Despite this difference in magnitude, which likely reflects differences in interaction modality and coordination infrastructure, both studies converge on the same directional pattern: as more AI partners join the group, humans redirect from direct content authorship toward steering, evaluation, and coordination. This convergence across studies with different tasks, interfaces, and AI counts suggests that the shift from co-author to orchestrator may be a general trajectory in human-AI collaboration, one that intensifies as AI partners multiply and coordination tools become available.

\textbf{Capture-stage shift.} The redirection from board authorship to conversation becomes visible in how much the human talks relative to how much they write. Participants in two-AI rooms continued to talk at comparable rates, but each human-authored board item was accompanied by more chat: 5.42 messages per board item compared with 2.52 in one-AI rooms (\(d\)=+0.88). In human-only brainstorming, the classic bottleneck is production blocking, where group members must wait to speak \cite{diehlStroebe1987}. In our human-AI groups, that bottleneck shifts from speaking to board capture: humans continue generating ideas in chat but write fewer of them to the shared board. Prior work has shown that proposing ideas and writing them to a shared artifact are separable stages \cite{mullerEtAl2024GroupBrainstorming, shaerEtAl2024Brainwriting}, and this rising conversion cost localizes the shift to the capture stage specifically.

The production-blocking account centers on a specific cognitive cost: when group members cannot verbalize ideas as they arise, those ideas may be forgotten or dismissed as less relevant by the time a turn opens \cite{diehlStroebe1987}. Our human-AI brainstorming eliminates this classical bottleneck entirely: participants typed freely in a parallel chat channel without waiting for turn-taking, yet human board authorship still declined. This finding extends the production-blocking literature by showing that removing the original bottleneck does not restore individual productivity when a new constraint takes its place. The new constraint is not temporal (waiting to speak) but ecological (the board is already populated). A complementary cognitive account holds that exposure to others' ideas functions as external stimulation that redirects the individual's memory search toward new semantic categories \cite{nijstadStroebe2006}. When AI items accumulate rapidly, they provide dense external cues that may accelerate category switching, but the board simultaneously fills with captured material, reducing the marginal value of recording one more idea to an artifact where much of the capture work is already done. This is not production blocking in the classical sense (the human is not prevented from speaking), but it is a capture-stage analog in which the cost of recording an idea to the shared artifact rises as the artifact fills.

\textbf{Formation and predictors.} The ecological constraint of a populated board explains the overall authorship decline, but the AI partner's behavioral profile determines how sharply it manifests. Tuning the AI's behavioral formation shapes human authorship with only one AI partner but ceases to matter with two AI partners. With one AVG AI, humans wrote more board items than in any other condition, including the human-human reference (25.7 vs 20.5, \(d\)=+1.60); with one CUS AI, they fell below it (17.7 vs 20.5, \(d\)=-0.62). With two AI partners, both formations converged near 13 items (\(d\)=-0.05, \(p\)=.862). Prior work on single-AI brainstorming has shown that a proactive agent's behavioral profile shapes how the human participates \cite{houdeEtAl2025Controlling}, consistent with formation mattering at one AI partner. Our OLS models suggest why formation ceases to matter with two AI partners: each additional AI-authored board item predicted approximately 0.27 fewer human board items, but AI social and filler talk predicted more (\(\beta\)=+26.11, \(p\)=.029), while generative, evaluative, and questioning AI messages were individually non-significant. These two predictors account for the formation boundary. With only one AI partner, the predictors diverge across formations: AVG devoted a substantial share of its talk to social conversation while producing fewer board items, whereas CUS produced minimal social talk and more board items. With two AI partners, both formations converge: board volume is high in both, and social talk collapses to near-zero in both, so the one predictor that differentiated formations disappears.

The convergence of formation effects at two AI partners could alternatively reflect social loafing: humans may reduce effort because the AI can be relied upon to produce sufficient output, particularly in the CUS condition where the AI was explicitly designed to reduce productivity blockers. However, the communication data argue against pure free-riding: humans in two-AI rooms maintained comparable chat rates and shifted toward generative messages rather than disengaging, a pattern more consistent with role redirection than motivational withdrawal. The mechanism is better characterized as declining marginal value of contribution: when the board fills quickly with AI-authored items, the perceived return from adding one more human item is similar whether the AI partner communicated as a conversational co-generator or as a coach.

\textbf{Social talk as scaffold.} Although accumulated board volume overwhelms formation differences at two AI partners, one dimension of the AI's communication profile does predict human authorship independently: social and filler talk. The positive association between AI social talk and human board authorship (\(\beta\)=+26.11, \(p\)=.029) stands in tension with the brainstorming literature, which has consistently treated social conversation as a productivity cost. The classical account characterized off-task social talk as detrimental, arguing that verbal discussion unrelated to the task consumes attention and working memory that would otherwise support idea generation \cite{paulus2007}. In that account, attention is a zero-sum resource, and time spent on social conventions is time diverted from idea generation. Our finding contradicts this characterization in human-AI brainstorming. When the AI devotes a larger share of its talk to low-pressure social conversation, the human writes more ideas to the board, not fewer. One explanation is that AI social talk serves a different cognitive function than human social talk. In human groups, social conversation is reciprocal: one person's off-task remark demands attention and response from others, consuming the shared attentional resource that model describes. AI social talk, by contrast, is non-reciprocal: the human can register a greeting or backchannel from the AI without needing to respond in kind, making it attentionally cheap. AI social talk may also function as a pacing signal that creates low-demand intervals between bursts of AI-generated ideas, giving the human time to process and commit their own ideas to the board. This interpretation is consistent with evidence that brief pauses in idea generation give individuals time to process what others have shared, renewing cognitive stimulation for the next burst of production \cite{paulus2007}. When AI social talk is absent (as in CUS rooms, where less than 1\% of AI messages were social), the AI produces a denser stream of substantive content that the human must continuously process, leaving fewer natural pauses for board capture. The result is that social talk, far from being a pure productivity cost, may serve as a temporal scaffold that sustains human authorship in AI-augmented brainstorming.

\textbf{Communication composition.} The lower human board count reflects a shift in what the human does within the team rather than disengagement or motivational withdrawal. Rather than talking less, humans changed what they talked about: they nearly eliminated social and filler talk (1 to 3\% vs 15\% in human-human groups) and redirected that capacity toward generative messages (69--71\% vs 58\% with two AI partners). This suppression of social talk is consistent with findings from a different collaborative domain, where human-AI teams sent more task-oriented messages and fewer interpersonal ones than human-human teams \cite{juAral2026Collaborating}. The two formations also created distinct interaction environments: with one AI partner, AVG agents operated as conversational co-generators (58\% generative, 22\% social filler), while CUS agents adopted a coaching profile (25\% evaluative, 32\% questioning, less than 1\% social filler) and authored more board items; with two AI partners, AVG agents shifted toward near-exclusive generation (77\% generative, 1.4\% social filler), narrowing the gap with CUS. The behavioral evidence is consistent with a role shift rather than disengagement: humans in AI conditions talked at comparable rates and redirected their social capacity toward idea generation rather than reducing their participation overall.

The suppression of human social talk in AI conditions is quantitatively consistent with findings from a large-scale field experiment on human-AI collaboration but reveals a different downstream consequence. In that study, human-AI teams sent 25\% more task-oriented messages focused on suggestions, instructions, and planning, while human-human teams sent 18\% more social and emotional messages focused on rapport building and self-assessment \cite{juAral2026Collaborating}. Our participants showed a comparable composition shift: social and filler talk dropped from 15\% in human-human groups to 1--3\% in AI conditions, and generative messages rose from 58\% to 69--71\% in two-AI rooms. Where our findings diverge from that study is in the consequence of this shift. In their study, the move toward task-oriented communication improved productivity in human-AI teams but interpersonal communication predicted higher output diversity, suggesting that the efficiency gains from task-oriented communication come at the cost of creative variation. We observe an analogous tension at the AI level: AI social talk predicted more human board items, yet human social talk nearly vanished. The human adapted by becoming maximally task-oriented even as the AI's social talk, when present, sustained their board authorship. This asymmetry suggests that the productive function of social conversation in human-AI teams may depend on directionality: AI-to-human social talk sustains engagement, while human-to-AI social talk is perceived as unnecessary and is suppressed.

\subsection{A second AI partner increases local differentiation more than global breadth}

Humans who write fewer items to the board write more locally distinctive ones. Turning to output quality (RQ2), any AI presence broadened the semantic range of human contributions, but adding a second AI partner did not extend this gain at the global level. Compared with the human-human reference, all AI conditions produced more globally diverse human-authored board items (\(d\)=+0.57 to +1.08). Adding a second AI partner, however, did not increase global diversity further (\(d\)=+0.25, \(p\)=.182). In contrast, local diversity, which compared each human item against only the other items within its own session, did increase with a second AI partner (\(d\)=+0.57, \(p\)=.002). This split mirrors a broader pattern: in a large-scale experiment, exposure to AI-generated examples increased how different ideas in the pool were from each other without making any individual idea more original \cite{ashkinazeEtAl2025AIIdeas}. Our results refine that observation by showing that the level of measurement matters. Unlike earlier brainstorming studies where external stimuli were kept separate from participant outputs \cite{wangEtAl2011Diversity}, our design gives AI partners full board access alongside the human, making the board a shared artifact rather than a human-only record. This means the reference ecology around the human's items changes as more AI items are added, which can produce local differentiation without a corresponding global effect. Seeing others' ideas tends to push people toward less-explored territory \cite{nijstadStroebe2006}, which would explain why human items stand out more locally when AI items fill in the common ground around them. By measuring diversity at both global and local reference levels, this study shows that a second AI partner increases semantic differentiation within the session even when it does not expand the broader idea space, a distinction that single-level measurement would miss.

The distinction between collective diversity and individual creativity has been established empirically: high AI exposure increased collective diversity without affecting individual creativity, making ideas in the pool more different from each other but not necessarily better \cite{ashkinazeEtAl2025AIIdeas}. Our global diversity result is consistent with the different-but-not-better pattern: all AI conditions exceeded the human-human reference on global diversity, but adding a second AI partner did not increase it further. Our local diversity result, however, goes beyond what their design could detect. Because that study tested passive exposure to AI ideas in sequential response chains rather than real-time co-authorship on a shared artifact, they measured diversity at a single level, across the full pool. Our two-level measurement reveals that the same AI presence that fails to expand the global idea space does increase how different each human idea is from the board around it. This is consistent with their dynamic finding that AI ideas partially reset convergence in sequential chains: in our shared-board setting, AI items continuously fill common semantic territory, pushing the human's next contribution toward less-explored ground within the session even when the study-wide idea space has already absorbed that territory.

The local–global split follows from how external cues redirect idea generation. Exposure to others' ideas can push individuals toward less-explored semantic categories and accelerate the rate at which they switch between categories \cite{nijstadStroebe2006}. In our setting, AI-authored board items function as a continuous stream of such cues. Whether these cues are diverse in the categorical sense the original experiments tested depends on the language model's output distribution, but they are certainly dense, arriving at a pace that exceeds what a human partner would produce. This stimulation accelerates the human's category switching within the session, pushing them toward semantic territory not yet covered on their room's board and producing local differentiation: the human's items become more distinct from the AI-populated board around them. However, because all sessions drew on the same underlying language model (GPT-4o), AI items across sessions occupy similar semantic territory. A human idea that is locally distinctive, far from the AI items on their own board, may still be globally common, similar to what humans in other rooms produced under similar AI stimulation. This account predicts exactly the split we observe: AI cues broaden within-session exploration (local effect) but do not escape the study-wide semantic envelope set by the language model's output distribution (no global effect).

Consistent with this account, behavioral formation changed how the AI communicated but not the semantic diversity of human-authored board ideas. Within a fixed group size, AVG and CUS formations produced comparable local diversity (1AVG vs 1CUS: \(d\)=-0.36, \(p\)=.164; 2AVG vs 2CUS: \(d\)=-0.15, \(p\)=.575). This null result is notable because deliberately diversified AI personas can preserve output diversity compared to a human-only baseline, and the relationship between generative AI and diversity may depend on how the AI is deployed \cite{wanKalman2026DiversePersonas}. That study varied AI personas at the content level to produce diverse outputs, yet our formations, which varied communication style rather than content, did not produce different diversity outcomes. One explanation is that the two idea sources, a curated bank of human-generated ideas for AVG and GPT-4o generation for CUS, cover a similar semantic range for this brainstorming topic, so varying the source did not produce measurably different diversity outcomes. This suggests that, for a constrained topic, the semantic territory available to brainstormers may be more strongly bounded by the topic itself than by whether ideas originate from a human corpus or a language model. This finding qualifies the design recommendation that AI should be treated as a tool that excels at delivering precisely what is specified rather than an intuitive collaborator that extracts unstated needs. In our case, the AVG formation was calibrated to match average human brainstorming behavior, and the CUS formation let participants configure process-loss dimensions whose emergent behavioral profile differed substantially, yet neither specification changed the semantic diversity of the ideas that reached the board.

\subsection{Implications}

Evaluating human-AI teams requires rethinking the metrics, design variables, and longitudinal assumptions that guide collaborative AI research. First, because adding AI partners as a design act can change what role the human plays rather than how much the human participates, evaluating human-AI teams on output volume alone will systematically undercount the human's contribution. Researchers and practitioners should measure conversation, composition, and authorship separately to capture the full picture of what each team member thinks and does. Second, the effects of AI team size on human contribution cannot be extrapolated from single-agent results: formation shaped human authorship with only one AI partner but not with two, meaning that design guidelines derived from one-agent studies may not transfer as teams scale. Third, the distinction between global and local diversity reveals that adding more AI partners can increase how different each human idea is from the board around it without broadening the overall idea space, a result that has practical consequences for how creative output is assessed in AI-augmented teams. More broadly, creativity research has long sought situational factors that augment what individuals produce beyond their solo capacity, yet decades of group brainstorming research consistently found that adding more people introduces process losses that offset any stimulation gain \cite{diehlStroebe1987, paulus1993}. Our findings suggest that AI partners can reverse this trajectory: a second AI increased within-session semantic differentiation of human ideas without imposing the coordination costs that make human groups less than the sum of their parts, indicating that the situational stimulus added by AI can function as a net positive rather than a net process cost.

The rate at which AI partners deliver content shapes human authorship as much as the content itself. The cognitive-social-motivational model of brainstorming positions attention as the central link between social factors and individual idea generation: attending to others' ideas can cognitively stimulate new combinations, but it diverts attentional resources from one's own idea generation \cite{paulus2007}. AI partners restructure this tradeoff. Because AI messages arrive as text that can be scanned rather than spoken turns that demand listening, the attentional cost of monitoring AI contributions may be lower than monitoring human partners. At the same time, the density and speed of AI output can overwhelm, as participants in a study of proactive AI agents reported that the AI produced text so quickly that it was hard to keep up \cite{houdeEtAl2025Controlling}. Our finding that AI social talk predicted higher human board authorship (as established in \S5.1) suggests a specific pacing mechanism. The brainstorming literature has treated social conversation as a pure cost to be minimized \cite{paulus2007}, and recent empirical work confirms that human-AI collaboration shifts communication toward task-oriented content \cite{juAral2026Collaborating}. Yet in our data, AI social talk created natural breaks between bursts of substantive content, functioning as attentionally lightweight intervals that gave the human time to process and commit their own ideas. When social talk was absent (as in CUS rooms, where less than 1\% of AI messages were social), the AI produced a denser stream of substantive content that the human must continuously process. The practical recommendation is that AI brainstorming systems should regulate the rate and density at which the AI contributes alongside what it says, treating information arrival rate as a first-class design variable. Calibrated social talk may serve as a lightweight pacing mechanism that sustains human authorship without requiring the human to invoke explicit throttling controls.

The authorship shift we observe in a single session may have consequences that extend beyond it. Recent evidence indicates that LLM assistance can provide short-term creativity boosts during assisted tasks but may hinder independent creative performance when users later work without assistance \cite{kumarEtAl2025HumanCreativity}. More concerning, participants who received LLM-generated strategies continued to generate more similar ideas even in later unassisted rounds, suggesting that the "homogenization effect" can persist after the AI is removed. Our study was not designed to measure post-session effects, but the sharp reduction in human board authorship with two AI partners, combined with the redirection of human communication toward generative chat that never reaches the board, raises the question of whether extended experience with AI-populated brainstorming boards could attenuate the human's habit of committing ideas to shared artifacts. If the capture-stage shift we document becomes habitual, it could affect the human's contribution patterns in future collaborations, with or without AI present. Future longitudinal work should examine whether the role shift we observe in a single session persists, reverses, or deepens across repeated human-AI brainstorming encounters.

\subsection{Limitations}

Several design choices scope these findings. The study tested a single brainstorming session using GPT-4o in a real-time text chat with a shared persistent board; however, the controlled 2$\times$2 crossing of group size and formation isolates variables that prior multi-agent studies bundled together, providing the first evidence that formation effects can vanish as AI team size grows. Participants interacted with AI partners in groups of two or three, and configurations with larger AI teams may produce qualitatively different dynamics; yet even at this scale, the interaction between group size and formation already reveals that single-agent design guidelines do not transfer directly to multi-agent settings.

\bibliographystyle{ACM-Reference-Format}
\bibliography{verified_references}

\appendix
\input{appendix}

\end{document}

%% file: appendix.tex
\section{Complete OLS Models Predicting Human Board Items}

\label{app:ols-models}

We estimated separate OLS regressions predicting human-authored board items per person, each entering one AI message-act category alongside AI board volume, group size, and formation as controls (\(N\)=120). Tables \ref{tab:app-ols-social}--\ref{tab:app-ols-question} report the full coefficient tables.

\subsection{Social/Filler Model}

\label{app:ols-social}

\begin{longtable}[]{@{}lrrrr@{}}
\caption{OLS model predicting human-authored board items with AI social/filler share as the focal predictor (\(R^2\)=.338). \label{tab:app-ols-social}}\tabularnewline
\toprule\noalign{}
Term & \(\beta\) & SE & \(t\) & \(p\) \\
\midrule\noalign{}
\endfirsthead
\toprule\noalign{}
Term & \(\beta\) & SE & \(t\) & \(p\) \\
\midrule\noalign{}
\endhead
\bottomrule\noalign{}
\endlastfoot
Intercept & 23.01 & 4.09 & +5.63 & \textbf{\textless.001***} \\
AI social/filler share & +26.11 & 11.81 & +2.21 & \textbf{.029*} \\
Group size (number of AI partners) & +0.56 & 2.57 & +0.22 & .827 \\
Formation (CUS vs.\ AVG) & +1.26 & 2.11 & +0.60 & .551 \\
AI-authored board items & -0.271 & 0.073 & -3.73 & \textbf{\textless.001***} \\
\end{longtable}

\subsection{Generative Model}

\label{app:ols-generative}

\begin{longtable}[]{@{}lrrrr@{}}
\caption{OLS model predicting human-authored board items with AI generative share as the focal predictor (\(R^2\)=.315). \label{tab:app-ols-generative}}\tabularnewline
\toprule\noalign{}
Term & \(\beta\) & SE & \(t\) & \(p\) \\
\midrule\noalign{}
\endfirsthead
\toprule\noalign{}
Term & \(\beta\) & SE & \(t\) & \(p\) \\
\midrule\noalign{}
\endhead
\bottomrule\noalign{}
\endlastfoot
Intercept & 33.61 & 4.16 & +8.08 & \textbf{\textless.001***} \\
AI generative share & -6.59 & 6.59 & -1.00 & .319 \\
Group size (number of AI partners) & -1.50 & 2.42 & -0.62 & .537 \\
Formation (CUS vs.\ AVG) & -3.23 & 2.16 & -1.50 & .137 \\
AI-authored board items & -0.272 & 0.075 & -3.61 & \textbf{\textless.001***} \\
\end{longtable}

\subsection{Evaluative Model}

\label{app:ols-evaluative}

\begin{longtable}[]{@{}lrrrr@{}}
\caption{OLS model predicting human-authored board items with AI evaluative share as the focal predictor (\(R^2\)=.317). \label{tab:app-ols-evaluative}}\tabularnewline
\toprule\noalign{}
Term & \(\beta\) & SE & \(t\) & \(p\) \\
\midrule\noalign{}
\endfirsthead
\toprule\noalign{}
Term & \(\beta\) & SE & \(t\) & \(p\) \\
\midrule\noalign{}
\endhead
\bottomrule\noalign{}
\endlastfoot
Intercept & 30.59 & 2.49 & +12.29 & \textbf{\textless.001***} \\
AI evaluative share & -12.04 & 10.49 & -1.15 & .254 \\
Group size (number of AI partners) & -2.21 & 2.21 & -1.00 & .319 \\
Formation (CUS vs.\ AVG) & +0.61 & 2.66 & +0.23 & .818 \\
AI-authored board items & -0.265 & 0.074 & -3.59 & \textbf{\textless.001***} \\
\end{longtable}

\clearpage

\subsection{Questioning Model}

\label{app:ols-question}

\begin{longtable}[]{@{}lrrrr@{}}
\caption{OLS model predicting human-authored board items with AI questioning share as the focal predictor (\(R^2\)=.311). \label{tab:app-ols-question}}\tabularnewline
\toprule\noalign{}
Term & \(\beta\) & SE & \(t\) & \(p\) \\
\midrule\noalign{}
\endfirsthead
\toprule\noalign{}
Term & \(\beta\) & SE & \(t\) & \(p\) \\
\midrule\noalign{}
\endhead
\bottomrule\noalign{}
\endlastfoot
Intercept & 29.80 & 2.70 & +11.04 & \textbf{\textless.001***} \\
AI questioning share & +2.49 & 5.61 & +0.44 & .658 \\
Group size (number of AI partners) & -2.36 & 2.23 & -1.06 & .292 \\
Formation (CUS vs.\ AVG) & -2.13 & 1.79 & -1.19 & .236 \\
AI-authored board items & -0.263 & 0.075 & -3.49 & \textbf{\textless.001***} \\
\end{longtable}

All coefficients are unstandardized. The reference category for formation is AVG; group size is coded as the number of AI partners (1 or 2). AI-authored board items is a count; message-act shares are proportions ranging from 0 to 1. Significance levels: *** \(p\)\textless.001, ** \(p\)\textless.01, * \(p\)\textless.05.